# Pendulum Model of Spiking Neurons

Joy Bose
*Ericsson*
Bangalore, India
joy.bose@ieee.org

*Abstract*— We propose a biologically inspired model of spiking neurons based on the dynamics of a damped, driven pendulum. Unlike traditional models such as the Leaky Integrate-and-Fire (LIF) neurons, the pendulum neuron incorporates second-order, nonlinear dynamics that naturally give rise to oscillatory behavior and phase-based spike encoding. This model captures richer temporal features and supports timing-sensitive computations critical for sequence processing and symbolic learning. We present an analysis of single-neuron dynamics and extend the model to multi-neuron layers governed by Spike-Timing Dependent Plasticity (STDP) learning rules. We demonstrate practical implementation with python code and with the Brian2 spiking neural simulator, and outline a methodology for deploying the model on neuromorphic hardware platforms, using an approximation of the second-order equations. This framework offers a foundation for developing energy-efficient neural systems for neuromorphic computing and sequential cognition tasks.

## I. Introduction

Spiking Neural Networks (SNNs) have emerged as a promising paradigm for temporal and event-driven processing, yet most models such as Izhikevich [1] or LIF [2] prioritize simplicity or biophysical realism. We propose a pendulum neuron model rooted in nonlinear dynamics, capable of capturing phase and rhythm in spiking behavior with relatively few parameters. Inspired by the mathematical pendulum, this model naturally accommodates temporal abstraction and symbolic spike timing, which are useful for sequence learning and timing-sensitive computations.

The pendulum model of spiking neurons presented in this paper is a biologically inspired evolution of the earlier wheel model introduced in the author's PhD thesis [11]. The wheel model was originally conceived as a symbolic and geometric abstraction for spike timing control, where a neuron's internal phase evolved akin to a rotating wheel, and spikes were emitted when the wheel crossed a designated angular threshold (e.g., $\theta=2\pi$). This model enabled precise spike encoding and was central to constructing sequence machines capable of recognizing and generating temporal patterns. However, while the wheel model was effective in modelling symbolic timing and deterministic sequence transitions, it lacked mechanisms for capturing the dynamic, continuous, and noisy nature of biological neurons. It also did not incorporate physical properties such as damping, resonance, or integration of external inputs in a realistic manner.

To address these limitations, the pendulum model generalizes the wheel by introducing a second-order nonlinear differential equation inspired by the dynamics of a damped driven pendulum. This formulation allows for a number of relevant features: oscillatory behaviour that naturally supports rhythmic spiking, phase-based temporal encoding aligned with biological oscillators, input-driven modulation of spike timing, and support for biologically plausible learning mechanisms such as Hebbian plasticity and STDP. Thus, the pendulum neuron can be seen as a biologically grounded continuation of the symbolic ideas embedded in the wheel model. While the wheel served as a conceptual foundation for discrete symbolic timing, the pendulum neuron bridges this with biological realism and is better suited for simulation and deployment on neuromorphic platforms. This lineage from the wheel to the pendulum preserves the key idea of phase-driven symbolic computation, while expanding it into the continuous-time, dynamic domain of spiking neural networks.

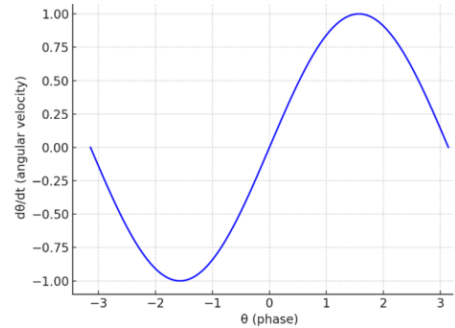

Fig. 1. An illustration of the phase space in the Pendulum spiking neuron model. The pendulum model introduces dynamic, second-order motion with biologically inspired thresholding

The original idea of wheel and pendulum model was inspired by earlier discussions of the author with physicist Mike Cumpstey and engineering a spiking sequence machine using a physical analogy to nonlinear oscillators.

## II. Pendulum Model Formulation

### A. Wheel Model equations

The wheel model [11], from which the pendulum model is inspired, represents a spiking neuron using uniform angular phase evolution on a circle. The neuron's internal state is the angular phase $\theta(t)$, which increases over time until it reaches a threshold, upon which the neuron fires a spike and resets. The basic form of the wheel model uses a first-order differential equation:

$$d\theta/dt = \omega \qquad (1)$$

Where:

$\theta(t) \in [0, 2\pi]$ is the angular phase of the neuron

$\omega$ is the constant angular velocity.

This results in uniform circular motion. Optionally, $\omega$ can be made time-varying to incorporate input:

$$d\theta/dt = \omega + \alpha\, I(t) \qquad (2)$$

Where: I(t) is the external input current and α is a scaling parameter.

### B. Pendulum Model Equation

This model is termed the "pendulum" model because its core dynamics are directly inspired by the physical behavior of a damped, driven pendulum. In classical mechanics, a pendulum exhibits oscillatory motion governed by second-order nonlinear differential equations, characterized by parameters such as damping, resonance, and external forcing. These features closely parallel the temporal dynamics observed in biological neurons, particularly those involved in rhythmic or timing-sensitive functions. By adopting this analogy, the pendulum model captures the rich temporal structure and phase-sensitive behavior that simpler first-order models such as the Leaky Integrate-and-Fire (LIF) cannot. Furthermore, the pendulum formulation naturally generalizes the wheel model, where spikes were emitted upon completing angular cycles, by embedding it within a more biologically realistic framework that includes inertia, energy dissipation, and continuous input modulation. Hence, the model derives its name both from its mathematical lineage and its functional resemblance to oscillatory neuronal circuits in the brain.

For the pendulum model, we define the neuron using a second-order differential equation:

$$\frac{\{d^2\theta\}}{\{dt^2\}} + \frac{\gamma\{d\theta\}}{\{dt\}} + \omega^2 \sin(\theta) = I(t) \qquad (3)$$

Where $\theta$ is the angular phase, $\gamma$ is damping, $\omega$ is the natural frequency, and I is the input current. A spike is emitted when $\theta \geq \pi$ after which the system resets to $\theta = 0$. This setup intrinsically models oscillatory, phase-aware behavior and provides richer temporal dynamics than the LIF model [2]. It is inspired by biological resonance and timing circuits, such as those of the cerebellum in the brain.

### C. Single Neuron Behavior

Simulations reveal that pendulum neurons spike periodically in response to constant input, with frequency determined by $\omega$ and input intensity I. Unlike LIF neurons [2], which integrate linearly toward a threshold, pendulum neurons can accelerate or decelerate based on both the current state and input, making them inherently non-linear.

### D. Hebbian Learning using Pendulum Neurons

The pendulum neuron model supports Hebbian plasticity [7, 10]. According to Hebb's classical principle [7] which mentioned that "neurons that fire together, wire together", a synapse is strengthened if the pre- and post-synaptic neurons spike within a short temporal window.

In the pendulum framework, this rule can be implemented by monitoring co-occurring spikes across neurons at each timestep. When two connected neurons emit spikes simultaneously or within a defined threshold window, the corresponding synaptic weight is incremented. This simple correlation-based learning rule enables the network to encode frequently co-active patterns and supports associative memory and feature detection.

### E. Layered Networks and STDP Learning

Spike-Timing Dependent Plasticity (STDP) [3, 6, 8] in pendulum neurons operates on the same core principle as in other spiking models: the precise timing of spikes determines the direction and magnitude of synaptic change. However, due to the oscillatory dynamics of the pendulum model, spike phases carry richer temporal context. This enables more nuanced learning rules that can associate temporal phase relationships with synaptic strength. For example, pre-synaptic spikes occurring slightly earlier than post-synaptic ones can lead to potentiation, while reversed timing results in depression, consistent with classical STDP profiles. Moreover, because the pendulum neuron's phase evolution is smooth and continuous, it supports interpolation between spike timings, allowing learning to be sensitive to small variations in temporal patterns. Such dynamics are especially beneficial for learning rhythmic sequences, where timing precision is essential.

We construct a layer of interconnected pendulum neurons with synapses modifiable via STDP [3, 6, 8]. The rule adjusts weights based on spike timing differences:

$$\Delta\omega = \begin{array}{l} A_+ e^{\left\{\frac{-\Delta t}{\tau_+}\right\}} \text{ if pre before post} \\ -A_- e^{\left\{\frac{\Delta t}{\tau_-}\right\}} \text{ if post before pre} \end{array} \qquad (4)$$

In networks, these neurons form oscillatory phase-locked patterns and can learn temporal associations in symbolic sequences, such as character streams or rhythmic patterns.

### III. COMPARISON OF PENDULUM MODEL WITH OTHER SPIKING MODELS

The pendulum neuron model offers an alternative to classic spiking neuron paradigms such as the Leaky Integrate-and-Fire (LIF) by emphasizing continuous-time, second-order dynamics. LIF neurons [2] are computationally efficient and widely used due to their simplicity, but they lack adaptive behavior and phase-sensitive dynamics. The Izhikevich model [1] strikes a balance between biological realism and computational cost, capturing rich firing patterns through carefully tuned parameters. In contrast, the pendulum neuron encodes spike timing via angular phase, enabling temporal abstraction and smoother interpolation of timing than both LIF and Izhikevich neurons. While it does not incorporate intrinsic adaptation mechanisms like the Izhikevich model, the pendulum neuron's inherent oscillatory behavior naturally aligns with periodic and sequence-based tasks. Its second-order dynamics introduce complexity in simulation but provide a richer feature space for learning and encoding symbolic patterns.

Table 1 shows a comparison of the pendulum model with other spiking neuron models: Leaky Integrate and Fire (LIF) and Izhikevich model.

TABLE I.  COMPARING PENDULUM MODEL WITH LIF AND IZHIKEVICH SPIKING NEURON MODELS

| Model | Order | Phase Encoding | Adaptation | Computation |
|---|---|---|---|---|
| LIF | First | No | No | Simple |
| Izhikevich | Second | Limited | Yes | Moderate |
| Pendulum | Second | Yes | No | Complex |

Pendulum neurons offer superior temporal abstraction but may require more computational resources. They are particularly effective in tasks involving timing, symbolic processing, or rhythm.

## IV. IMPLEMENTATION IN PYTHON FOR SINGLE PENDULUM NEURONS AND STDP LEARNING RULE

### A. Python Implementation for a single neuron

Below is Python code to implement a Pendulum spiking neuron subject to a time varying input current I(t).

```
# Parameters
T = 500            # Total time (ms)
dt = 0.1           # Time step (ms)
steps = int(T/dt)
time = np.linspace(0, T, steps)

# Pendulum parameters
gamma = 0.05       # Damping coefficient
omega = 1.0        # Natural frequency
theta_reset = 0.0  # Reset value after spike
threshold = np.pi  # Threshold for spike

# Input current
def input_current(t):
    return 1.5 * np.sin(0.01 * t) + 1.2  # Example of sine + bias

# Initialize state variables
theta = np.zeros(steps)
dtheta = np.zeros(steps)
spikes = np.zeros(steps)

# Euler integration
for i in range(1, steps):
    I = input_current(time[i])
    ddtheta = -gamma * dtheta[i-1] - omega**2 * np.sin(theta[i-1]) + I
    dtheta[i] = dtheta[i-1] + ddtheta * dt
    theta[i] = theta[i-1] + dtheta[i] * dt

    # Check for spike
    if theta[i] >= threshold:
        spikes[i] = 1
        theta[i] = theta_reset
        dtheta[i] = 0.0
```

The output of the Python code, showing the pendulum neuron dynamics for a single neuron, is shown in figure 2.

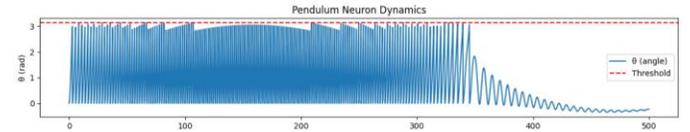

Fig. 2. Pendulum neuron dynamics for a single neuron.

### B. Pseudocode for a layer of multiple pendulum neurons (without learning)

Below is pseudocode for a layer of pendulum spiking neurons, with an external input current I.

```
Initialize:
  N pendulum neurons
  For each neuron i:
    theta[i]  ← 0         // angular position
    omega[i]  ← 0         // angular velocity
    I[i] ← external or input current
  Parameters:
    gamma    ← damping factor
    omega0   ← natural frequency
    dt       ← simulation time step
    threshold ← π
  Spike_log[i] ← empty list for storing spike times

For time t = 0 to T_max with step dt:

  For each neuron i = 1 to N:
    // Update pendulum dynamics
```

```
    domega = -gamma * omega[i] - omega0^2 *
sin(theta[i]) + I[i]
    omega[i] += domega * dt
    theta[i] += omega[i] * dt

    // Check for spike
    If theta[i] ≥ threshold:
        Append t to Spike_log[i]
        Reset: theta[i] = 0, omega[i] = 0

Repeat until t reaches T_max
```

### C. Pseudocode of STDP Learning Algorithm for Pendulum Neurons

The pseudocode for implementing the STDP (Spike Time Dependent Plasticity) algorithm [3, 6, 8], a popular Hebbian style learning algorithm, for pendulum neurons is given below.

```
Initialize:
    N neurons with theta = 0, omega = 0
    Synaptic weights W[N][N] initialized randomly or to zero
    Time constants: tau_plus, tau_minus
    Learning rates: A_plus, A_minus
    Spike threshold: theta_thresh = pi
    Time step: dt

For each simulation step t:
    For each neuron i:
        Compute:
            dtheta_i = omega_i * dt
            domega_i = (-gamma * omega_i - omega0^2 *
sin(theta_i) + I_i) * dt

        Update:
            theta_i += dtheta_i
            omega_i += domega_i

        If theta_i >= theta_thresh:  // neuron i spikes
            Record spike time t_i
            Reset: theta_i = 0, omega_i = 0

            For all neurons j ≠ i:
                If neuron j spiked at t_j:
                    Δt = t_i - t_j
                    If Δt > 0:   // pre before post
                        W[i][j] += A_plus * exp(-Δt / tau_plus)
                    Else:        // post before pre
                        W[i][j] -= A_minus * exp(Δt / tau_minus)

    Update all neuron inputs I_i based on W and previous spikes
    Repeat until end of simulation
```

### V. IMPLEMENTATION IN BRIAN 2 SPIKING NEURON SIMULATOR

We implement the pendulum neuron equation in Brian 2 [4], a popular spiking neuron simulator, using custom differential equations:

```
from brian2 import *
eqs = '''
dtheta/dt = omega : 1
domega/dt = -gamma * omega - omega0**2 * sin(theta) + I : 1
I : 1
'''
G = NeuronGroup(1, model=eqs, threshold='theta > pi',
reset='theta=0; omega=0', method='euler')
```

This framework allows for rapid prototyping and scaling to multi-neuron systems using Brian 2 simulator [4].

In addition to Brian 2, we also explore implementation on neuromorphic hardware platforms such as SpiNNaker for real-time spiking simulation, in the following section.

### VI. IMPLEMENTATION OF PENDULUM NEURONS IN SPINNAKER NEUROMIMETRIC HARDWARE

SpiNNaker (Spiking Neural Network Architecture) [5] is a massively parallel, asynchronous neural simulation platform developed at the University of Manchester, capable of modeling large-scale SNNs in real time. Although it is optimized for models like LIF, custom neuron models such as the pendulum neuron can be approximated using hybrid approaches.

The pendulum neuron model can be implemented on the SpiNNaker neuromorphic platform [5] using the PyNN programming interface. Another way is to use sPyNNaker which supports custom neuron models in C, but only 1st-order ODEs natively. The template is given in [13].

Since SpiNNaker supports only first-order differential equations and fixed-point arithmetic, the model's second-order dynamics must be reformulated into coupled first-order updates. The nonlinear sine term can be approximated using precomputed lookup tables. Custom C code can define the neuron state updates, while PyNN can manage network configuration and simulation flow. This approach offers a viable path to deploying phase-encoded neural dynamics on low-power, event-driven hardware, not only for Spinnaker but also for other hardware implementations such as Loihi [9].

## VII. Conclusion

In this paper, we have introduced the pendulum model of spiking neurons. The pendulum neuron model provides a compelling alternative to traditional spiking models by incorporating temporal dynamics, phase encoding, and rich nonlinear behavior. Its ease of implementation in tools like Brian2 and compatibility with neuromorphic hardware make it a strong candidate for future bio-inspired computing architectures.

Code accompanying the simulations, including the Brian2 implementation and example STDP network, is available [12].

Future work includes integrating with rank-order encodings, symbolic memory, and benchmarking on real-world sequential tasks.